\documentclass[runningheads]{llncs}
\usepackage[T1]{fontenc}
\usepackage{graphicx}
\usepackage[english]{babel}
\usepackage{hyphenat}
\usepackage{booktabs}
\usepackage{amssymb}

\usepackage[verbose]{hyperref}

\begin{document}
\title{Evaluating the Effectiveness of Large Language Models in Automated News Article Summarization}
\titlerunning{Evaluating Large Language Models for News Summarization}
\author{Lionel Richy Panlap Houamegni\inst{1} \and Fatih Gedikli\inst{1}\orcidID{0000-0001-6190-0449}}
\authorrunning{Panlap Houamegni and Gedikli}
\institute{Institute of Computer Science \\
University of Applied Sciences Ruhr West \\
Mülheim an der Ruhr, Germany \\
\email{lionel.panlaphouamegni@stud.hs-ruhrwest.de, fatih.gedikli@hs-ruhrwest.de}
}
\maketitle              % typeset the header of the contribution
\begin{abstract}
The automation of news analysis and summarization presents a promising solution to the challenge of processing and analyzing vast amounts of information prevalent in today's information society. Large Language Models (LLMs) have demonstrated the capability to transform vast amounts of textual data into concise and easily comprehensible summaries, offering an effective solution to the problem of information overload and providing users with a quick overview of relevant information. A particularly significant application of this technology lies in supply chain risk analysis. Companies must monitor the news about their suppliers and respond to incidents for several critical reasons, including compliance with laws and regulations, risk management, and maintaining supply chain resilience. This paper develops an automated news summarization system for supply chain risk analysis using LLMs. The proposed solution aggregates news from various sources, summarizes them using LLMs, and presents the condensed information to users in a clear and concise format. This approach enables companies to optimize their information processing and make informed decisions. Our study addresses two main research questions: (1) Are LLMs effective in automating news summarization, particularly in the context of supply chain risk analysis? (2) How effective are various LLMs in terms of readability, duplicate detection, and risk identification in their summarization quality? In this paper, we conducted an offline study using a range of publicly available LLMs at the time and complemented it with a user study focused on the top performing systems of the offline experiments to evaluate their effectiveness further. Our results demonstrate that LLMs, particularly Few-Shot GPT-4o mini, offer significant improvements in summary quality and risk identification.

\end{abstract}

\section{Introduction}

In an increasingly interconnected and globalized economy, companies must efficiently manage complex supply chains while identifying potential risks early. Recent events, such as the COVID-19 pandemic and geopolitical tensions, have highlighted the vulnerabilities of global supply chains and underscored the need for proactive risk analysis \cite{Ivanov18062021,Hohenstein2022}. In this context, automated processing and analysis of news information are becoming essential to help companies identify and assess supply chain risks in real time. Automated news summarization helps reduce information overload, speed decision making, and improve awareness of important events. It can identify global events and trends that affect the supply chain, highlight early warning signals for potential risks, and improve supplier and market monitoring through continuous news analysis.

This study explores how modern LLMs can revolutionize automated news summarization for supply chain risk analysis. LLMs such as GPT-4, Mistral, and LLaMA have made remarkable advancements in natural language processing, opening new possibilities for efficiently analyzing and summarizing large volumes of text.

The objective of this research is to evaluate and compare the performance of various LLMs in the context of automated news summarization for supply chain risk analysis. Both quantitative and qualitative metrics are used to provide a comprehensive evaluation of model performance. A key focus is on various prompting techniques, including zero-shot, few-shot, and fine-tuning approaches, to evaluate their impact on summary quality.

The significance of this research arises from the growing demand for efficient and reliable risk analysis methods in supply chains \cite{Holgado2024,Ivanov18062021}. Integrating LLMs into existing risk management frameworks could enable companies to respond more quickly to potential disruptions and enhance supply chain resilience. Furthermore, this work contributes to the broader scientific discourse on the applicability and limitations of LLMs in domain-specific tasks.

Methodologically, this study employs an experimental design in which various LLMs are evaluated using a carefully curated dataset of news articles. The assessment utilizes established metrics such as ROUGE, BLEU, and BERTScore, alongside qualitative analyses leveraging G-Eval and human evaluators. This combination of quantitative and qualitative methods provides a nuanced understanding of model performance, considering factors such as accuracy, efficiency, scalability, and cost.

Research on news summarization and its applications in risk analysis remains an evolving field with ongoing challenges. Key issues, including factual accuracy, the tendency of models to hallucinate, and inherent biases, remain at the forefront of current discussions and constitute essential aspects of this study.

The findings of this investigation aim not only to advance research in AI and supply chain management, but also to provide practical insights for companies looking to integrate LLMs into their risk analysis workflows.

\section{Related Work}

The field of automated text summarization, particularly in the context of supply chain risk analysis, has evolved significantly with the advent of LLMs. This section provides an overview of prior research, highlighting key advancements, methodologies, and persisting challenges.

\subsection{Advancements in Automated News Summarization}

Liu and Lapata (2019) demonstrated that pretrained transformer models offer substantial improvements in news summarization. By integrating extractive and abstractive techniques, they improved summary coherence and informativeness, particularly when fine-tuned for domain-specific applications \cite{Liu2019}.

Stiennon et al. (2020) explored human feedback-driven training for summarization. They developed a reward model based on human preference alignment, which significantly improved model performance compared to traditional supervised learning techniques \cite{Stiennon2020}.

Goyal et al. (2022) assessed GPT-3's capabilities in news summarization. They found that GPT-3 could generate high-quality summaries without explicit fine-tuning, often outperforming human-generated summaries in coherence and fluency \cite{Goyal2022}.

Wei et al. (2022) introduced chain-of-thought prompting, demonstrating how structured reasoning prompts enable LLMs to perform complex summarization tasks with greater contextual awareness \cite{wei2022chain}.

Liu et al. (2023) proposed using LLM-generated summaries as reference training data for smaller models. Their approach enhanced the efficiency of smaller-scale summarization models, presenting a cost-effective alternative to fine-tuning large models \cite{liu2023learning}.

Zhang et al. (2024) studied the impact of instruction tuning on zero-shot summarization. They found that aligning models with high-quality reference summaries significantly improved performance, often rivaling human-written summaries \cite{zhang2024benchmarking}.

Further, Zhang et al. (2024) provided a systematic review of text summarization. They examined the transition from extractive to abstractive techniques and highlighted key challenges, such as factual accuracy, handling long documents, and mitigating biases in generated content \cite{zhang2024systematic}.

An underexplored aspect of summarization is identifying related news articles—also known as news story chains. Gedikli et al. (2021) addressed this challenge in \cite{Gedikli2021} by leveraging clustering and Named Entity Recognition (NER) to create datasets for automated story chain detection, significantly reducing manual labeling efforts while maintaining high-quality outputs. Similarly, Stockem Novo and Gedikli (2023) \cite{StockemNovo2023} investigate BERT-based methods for near-duplicate news article detection, highlighting the importance of NER in identifying duplicate content. Accurate duplicate detection is essential for risk news summarization, as redundant information can distort risk assessments and lead to inefficiencies in decision-making.

\subsection{Research Gaps and Contributions of This Work}

Despite significant progress, several challenges remain unaddressed in the field of automated news summarization, particularly in the context of supply chain risk analysis. This study aims to bridge the following gaps:

\begin{itemize}
  \item \textbf{Domain-Specific Adaptation:} Most existing LLMs are not optimized for supply chain risk analysis. This study explores methods to tailor LLMs for industry-specific summarization.
  \item \textbf{Factual Accuracy and Bias Mitigation:} Ensuring the reliability of generated summaries remains a critical issue. We investigate strategies to enhance factual accuracy and mitigate biases \cite{zhang2024benchmarking}.
  \item \textbf{Real-Time Integration:} The deployment of LLMs for real-time news analysis and early risk detection has been insufficiently explored. This work examines the feasibility of integrating LLMs into dynamic monitoring systems.
  \item \textbf{Evaluation of Readability and Duplicate Detection:} While some models produce coherent summaries, their effectiveness in detecting duplicate information and maintaining readability in large-scale analysis remains unclear.
  \item \textbf{Fine-Tuning for Risk Identification:} There is a lack of research on optimizing LLMs specifically for identifying and categorizing risk-related information in news articles relevant to supply chain disruptions.
\end{itemize}

By addressing these gaps, our work contributes to the ongoing development of AI-driven risk analysis solutions, demonstrating how LLMs can be effectively leveraged to enhance supply chain monitoring and resilience.

\section{Methodology}

This study adopts a mixed-methods approach, combining qualitative and quantitative techniques to ensure a comprehensive and nuanced analysis. Data will be collected from a diverse range of news sources, including local, regional, and international news agencies, as well as digital magazines. When news articles were not originally in English, they were translated using state-of-the-art translation models from Hugging Face\footnote{\url{https://huggingface.co}} to ensure consistency in analysis. However, due to copyright restrictions, the original news articles used in this study cannot be publicly shared.

To develop the news summarization system, a variety of large language models (LLMs) will be employed, including GPT-4o, GPT-4o mini, GPT-3.5 Turbo, Mistral Large 2, Mistral 8x22b, Mistral 7b, Llama-3.2 90b, Llama-3.1 70b, Llama 3.1 8b, Llama 3 70b, Llama 3 8b, Gemma 2-9b, Gemma 7b, and two fine-tuned models (one based on GPT-4o mini and another on GPT-3.5 Turbo). These models will be evaluated under different learning paradigms—zero-shot learning, few-shot learning, and fine-tuning—to optimize their performance and the accuracy of the generated summaries.

\subsection{Experimental Design}

\begin{itemize}
    \item \textbf{Input:} News articles from local, regional, and international sources, including digital magazines, translated to English when needed.
    \item \textbf{Output}: Summarized, duplication-free news articles that highlight potential risks in supply chains, enabling informed decision-making.
    \item \textbf{Model Comparison:} A thorough comparison of the performance of various LLMs will be conducted. The evaluation will focus on the quality of the summaries and the models' ability to identify risks. Different learning strategies—zero-shot learning, few-shot learning, and fine-tuning—will be employed to optimize the models' effectiveness.
    \item \textbf{Hypothesis:} The latest LLMs will produce higher-quality summaries and more accurate risk identification compared to older models.
    \item \textbf{Independent Variable:} The specific language model used.
    \item \textbf{Dependent Variables:} The quality of the summaries and the accuracy of risk identification.
    \item \textbf{Procedure:} News articles will be input into the various language models, and the resulting summaries will be evaluated based on precision, relevance, and the ability to identify risks. Additional metrics such as price, speed, latency, and context windows will also be considered. The results will be subjected to statistical analysis to compare the performance of the models.
    \item \textbf{Expected Results:} It is anticipated that newer models, such as GPT-4o for proprietary models and LLaMA-3.2 90b for open-source models, will deliver superior summary quality and more precise risk identification.\footnote{See \url{https://artificialanalysis.ai} for a broad comparison of AI models and API providers.} This improvement is expected to provide a stronger foundation for decision-making in supply chain management.

    It is important to emphasize that these models represent the state of the art in LLMs at the time of our study (October 2024).
\end{itemize}

\subsection{Workflow}

\begin{figure}
    \centering
    \includegraphics[width=1\linewidth]{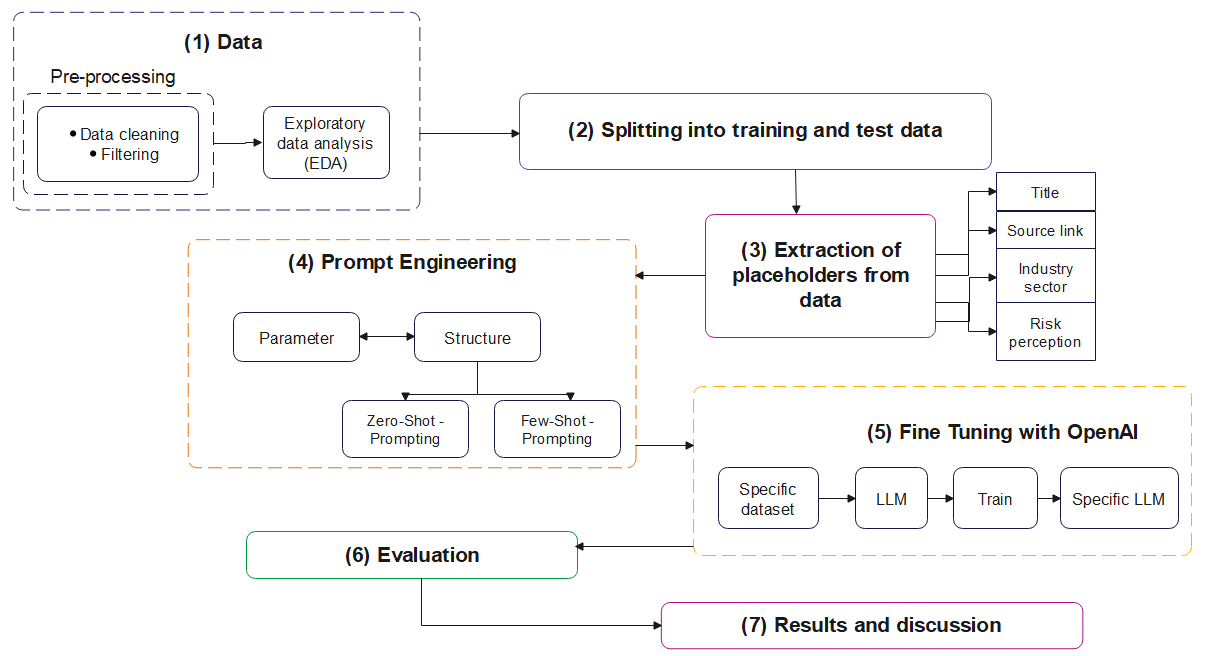}
    \caption{Evaluation Workflow}
    \label{fig:workflow}
\end{figure}

The workflow for this study is designed to systematically address the challenges of supply chain risk analysis using computational intelligence techniques. The process is divided into seven key stages, as illustrated in Figure \ref{fig:workflow}, and described below:

\textbf{(1) Data Import and Exploratory Data Analysis (EDA):}  

The raw dataset consists of \textbf{1,535} press articles, each enriched with metadata, including URLs, publication dates, headlines, abstracts, and sentiment ratings. Additional attributes capture industry classifications, risk categories, and references to key entities such as people, organizations, and locations. The dataset, provided by \textbf{graphworks.ai}\footnote{\url{https://graphworks.ai}}—an AI-powered platform for real-time news monitoring—spans \textbf{28 industries}, with \textit{Industrial Manufacturing and Services}, \textit{Transportation}, and \textit{Energy} being the most prevalent. This distribution aligns with key focus areas in supply chain risk analysis. Risk classifications encompass broad categories such as \textit{Interruption and Shutdown}, as well as more specific concerns like \textit{Cybersecurity and Data Privacy}. Initial data cleaning involved removing irrelevant columns, correcting inconsistent values, and ensuring data integrity for downstream tasks.

\textbf{(2) Separation of Training and Test Data:}  
To ensure a robust evaluation, the dataset was divided into distinct training and test subsets using an \textbf{80/20 split}. Specifically, \textbf{1,228} articles (\textbf{80\%}) were allocated for training, while \textbf{307} articles (\textbf{20\%}) were reserved for testing. This division ensures a balanced evaluation of model performance, allowing for an unbiased assessment of generalizability on unseen data.

\textbf{(3) Metadata Extraction:}
A computational pipeline is implemented to extract metadata, including article titles, original links, industry classifications, and risk perceptions. This metadata is integrated into the input parameters for summarization models to enhance contextual understanding and mitigate hallucinations, ensuring consistency across different LLMs.

\textbf{(4) Prompt Engineering:}
Effective prompt design is crucial for guiding LLMs in generating high-quality summaries. Drawing on best practices in prompt engineering\footnote{OpenAI, Prompt Engineering - Enhance Results with Prompt Engineering Strategies. \url{https://platform.openai.com/docs/guides/prompt-engineering}}, we employ few-shot prompting techniques \cite{mann2020language} to improve performance. Key parameters are carefully configured:
\begin{itemize}
\item \textbf{Maximum Token Count}: 4096 (total length of input and output)
\item \textbf{Temperature}: 0.3 (optimized through iterative experimentation)
\item \textbf{Top P (nucleus sampling)}: 0.5 (determined via systematic testing)
\end{itemize}

\textbf{(5) Fine-Tuning:}  
In addition to evaluating the base models, we fine-tune GPT-4o mini and GPT-3.5 Turbo for the specific task of news summarization in supply chain risk analysis using our domain-specific dataset. This step leverages computation techniques such as LoRA \cite{hu2022lora} to adapt the models to the unique characteristics of the dataset, enhancing their ability to capture nuanced risk-related information. Both fine-tuned and non-fine-tuned variants are tested to compare their performance.  

It should be noted that due to cost, time, and resource constraints, we could not utilize the entire training dataset during fine-tuning. Additionally, OpenAI's fine-tuning framework imposes a token limit of \(\leq 4096\) tokens per training example. As a result, in some cases, not all 10 articles for a given summary fit within the context window, requiring truncation to meet the token limit constraint. However, each training example contained at least 7–10 news articles along with a corresponding summary.

\textbf{(6) Evaluation:}
A comprehensive evaluation framework is employed to compare model performance across zero-shot, few-shot, and fine-tuned settings. Key evaluation criteria include:
\begin{itemize}
\item \textbf{Summary Quality}: Assessed using metrics such as coherence, relevance, and informativeness.
\item \textbf{Computational Efficiency}: Measured in terms of cost (USD per million tokens), output speed (tokens per second), latency (time to first token), and context window size.
\item \textbf{G-Eval}: A novel evaluation framework that provides deeper insights into model strengths and weaknesses, enhancing the accuracy of performance assessments.This method follows best practices as described by Anadkat and Fishman.\footnote{Shyamal Anadkat and Simón Fishman. How to evaluate a summarization task. \url{https://cookbook.openai.com/examples/evaluation/how_to_eval_abstractive_summarization}}
\item \textbf{Human Evaluation}: Essential for validating machine-generated summaries, human evaluation ensures alignment with domain-specific requirements and user expectations \cite{reiter2009investigation,liu2022revisiting,schuff2023human}.
\end{itemize}

\textbf{(7) Results and Discussion:}
The final results are presented, showcasing the effectiveness of the proposed workflow in generating high-quality summaries for supply chain risk analysis. The findings are analyzed in detail, with a discussion on their implications and potential applications. Additionally, future research directions are identified, particularly the integration of advanced LLM-based approaches (e.g., retrieval-augmented generation, reinforcement learning with human feedback) to further enhance model performance.

\section{Results}
\subsection{Inference with different LLMs}

This section presents the experimental results of employing multiple LLMs to generate summaries for news articles. In the context of LLMs, inference refers to the process of utilizing a trained model to generate predictions or derive insights from new, unseen data. Inference can be conducted using various approaches, including zero-shot, few-shot, and fine-tuning. This study evaluates the performance of several state-of-the-art language models\footnote{GPT-4o, GPT-4o mini, GPT-4o mini Fine-Tuned, GPT-3.5 Turbo, GPT-3.5 Turbo Fine-Tuned, Mistral Large 2, Mistral 8x22b, Mistral 7b, Llama-3.2 90b, Llama-3.1 70b, Llama 3.1 8b, Llama 3 70b, Llama 3 8b, Gemma 2-9b, and Gemma 7b. These models represent the state of the art available at the time the experiments were conducted.} to address the central research question: \emph{Can zero-shot prompting, few-shot prompting, and fine-tuning provide effective solutions for summarizing news articles in the context of supply chain risk analysis?}

In light of recent research findings demonstrating that few-shot learning consistently outperforms zero-shot approaches and significantly enhances model performance \cite{mann2020language}, GPT-4o is employed as the reference model in a few-shot learning context. As part of the latest GPT-4 generation, GPT-4o strikes an optimal balance between performance and accessibility, making it particularly suitable for research applications. The model has demonstrated exceptional capabilities in complex language understanding tasks, including reasoning and knowledge-based benchmarks such as the Massive Multitask Language Understanding (MMLU) \cite{hendrycks2021measuring}, MMLU-Pro \cite{nips_WangMZNCGRAHJLK24}, and General Purpose Question Answering (GPQA) \cite{rein2024gpqa}.
These achievements underscore GPT-4o's robustness and versatility, solidifying its position as a reliable choice for this study. Additionally, a human expert has reviewed and modified the GPT-4o reference summary where necessary, ensuring the accuracy and comprehensiveness of the information presented.

\subsection{Quantitative Evaluation}
\subsubsection{Automatic Analysis Using Similarity Metrics}

\begin{table}[htbp]
\caption{Evaluation Metrics Using Reference Summaries (Sorted by Average)}
\vspace{3mm}
\centering
\begin{tabular}{@{}ccccccc@{}}
\toprule
\textbf{Model} & \textbf{R-1} & \textbf{R-2} & \textbf{R-L} & \textbf{BLEU} & \textbf{BERTScore} & $\varnothing$ \\
\midrule
Zero-Shot GPT-4o & 0.781 & 0.601 & 0.774 & 0.595 & 0.950 & 0.7402 \\
Zero-Shot Mistral Large 2 & 0.719 & 0.494 & 0.701 & 0.449 & 0.941 & 0.6608 \\
Few-Shot GPT-4o mini & 0.734 & 0.480 & 0.720 & 0.389 & 0.935 & 0.6516 \\
Zero-Shot GPT-3.5 & 0.708 & 0.469 & 0.703 & 0.406 & 0.932 & 0.6436 \\
Few-Shot LLaMA3 70B & 0.717 & 0.464 & 0.706 & 0.330 & 0.948 & 0.6330 \\
Few-Shot Mistral Large 2 & 0.676 & 0.464 & 0.663 & 0.398 & 0.945 & 0.6292 \\
Fine-Tuned GPT-4o mini & 0.677 & 0.448 & 0.659 & 0.374 & 0.934 & 0.6184 \\
Zero-Shot LLaMA3 70B & 0.662 & 0.422 & 0.652 & 0.376 & 0.920 & 0.6064 \\
Zero-Shot GPT-4o mini & 0.683 & 0.411 & 0.660 & 0.354 & 0.918 & 0.6052 \\
Few-Shot Gemma2 9B & 0.649 & 0.409 & 0.624 & 0.374 & 0.945 & 0.6002 \\
Few-Shot GPT-3.5 & 0.682 & 0.371 & 0.662 & 0.316 & 0.943 & 0.5948 \\
Fine-Tuned with GPT-3.5 & 0.645 & 0.403 & 0.631 & 0.351 & 0.923 & 0.5906 \\
Few-Shot LLaMA3-1 70B & 0.645 & 0.409 & 0.629 & 0.335 & 0.935 & 0.5906 \\
Few-Shot LLaMA3-2 90B & 0.659 & 0.398 & 0.648 & 0.283 & 0.922 & 0.5820 \\
Few-Shot LLaMA3-1 8B & 0.631 & 0.401 & 0.618 & 0.316 & 0.939 & 0.5810 \\
Zero-Shot Mistral 7B & 0.622 & 0.385 & 0.601 & 0.356 & 0.925 & 0.5778 \\
Zero-Shot Mistral 822B & 0.638 & 0.390 & 0.626 & 0.309 & 0.910 & 0.5746 \\
Few-Shot Mistral 7B & 0.649 & 0.403 & 0.627 & 0.289 & 0.903 & 0.5742 \\
Zero-Shot Gemma2 9B & 0.636 & 0.394 & 0.627 & 0.305 & 0.908 & 0.5740 \\
Few-Shot LLaMA3 8B & 0.632 & 0.395 & 0.613 & 0.295 & 0.934 & 0.5738 \\
Zero-Shot LLaMA3 8B & 0.632 & 0.391 & 0.621 & 0.264 & 0.924 & 0.5664 \\
Zero-Shot LLaMA3-1 8B & 0.626 & 0.383 & 0.615 & 0.276 & 0.915 & 0.5630 \\
Zero-Shot LLaMA3-2 90B & 0.626 & 0.368 & 0.611 & 0.293 & 0.903 & 0.5602 \\
Zero-Shot LLaMA3-1 70B & 0.624 & 0.366 & 0.611 & 0.290 & 0.906 & 0.5594 \\
Few-Shot Mistral 822B & 0.577 & 0.330 & 0.559 & 0.273 & 0.905 & 0.5288 \\
Zero-Shot Gemma 7B & 0.476 & 0.225 & 0.454 & 0.163 & 0.877 & 0.4390 \\
Few-Shot Gemma 7B & 0.227 & 0.059 & 0.197 & 0.004 & 0.813 & 0.2600 \\
\bottomrule
\end{tabular}
\label{tab:performance}
\end{table}

This analysis \ref{tab:performance} compares various LLMs under different configurations (zero-shot, few-shot, fine-tuning). Note that R-1, R-2, and R-L represent ROUGE-1, ROUGE-2, and ROUGE-L scores, respectively. Tests were conducted by having each LLM summarize 10 articles in each summarization step. Key findings from the quantitative evaluation of automated summaries using similarity metrics include:
\begin{itemize}
    \item \textbf{Model Performance}: GPT-4o (zero-shot) achieved the best performance with an average score of 0.7402 across all metrics. Mistral Large 2 (zero-shot) and GPT-4o mini (few-shot) followed in 2nd and 3rd place.
    \item \textbf{Comparison of Prompting Methods}: Zero-shot prompting showed surprisingly good results, especially with advanced models like GPT-4o and Mistral Large 2. Few-shot prompting improved performance for some models but was not consistently superior. Fine-tuning (e.g., with GPT-4o mini) yielded strong results but did not achieve top performance.
    \item \textbf{Model Sizes and Performance}: Larger models (e.g., GPT-4o, Mistral Large 2) tended to perform better, while smaller models like Gemma 7B showed significantly weaker performance.
    \item \textbf{Metrics Comparison}: BERTScore consistently showed high values, indicating good semantic similarity. Similarly, ROUGE-1 and ROUGE-L provided consistent results for evaluating summary quality.
\end{itemize}

\subsubsection{Multi-criteria comparative analysis}

The multi-criteria comparative analysis includes metrics such as context window, output speed, latency, costs, and token processing. Key findings include:
\begin{itemize}
    \item \textbf{Summary Quality and Context Window}: Models like GPT-4o, GPT-4o mini, LLaMA 3.1 (70B), LLaMA 3.2, and Mistral Large 2 achieve outstanding results in summarization and risk identification. They offer a large context window of 128,000 tokens, enabling detailed analyses.
\end{itemize}
\begin{itemize}
    \item \textbf{Efficiency in Speed and Cost}: LLaMA models, especially LLaMA3-8B, are leaders in token processing rate, offering an ideal combination of efficiency and value for money. Open-source models like Gemma 2, Mistral 7B, Mistral 8x22B, and the LLaMA series generally have lower costs per token.
\end{itemize}
\begin{itemize}
    \item \textbf{Prompting Methods}: Few-shot prompting proved more effective for many models, especially in complex tasks. However, zero-shot prompting showed surprisingly strong results with models like GPT-4o, Mistral Large 2, and LLaMA 3 70B.
\end{itemize}
\begin{itemize}
    \item \textbf{Cost Analysis and Efficiency}: Proprietary models like GPT-4o and Mistral Large 2 guarantee high quality but are more cost-intensive. Open-source models (e.g., LLaMA, Gemma, Mistral 8x22B, and Mistral 7B) offer more cost-effective token and processing times.
\end{itemize}
\begin{itemize}
    \item \textbf{Recommended Models and Configurations}: Models with large context windows (e.g., GPT-4o, GPT4o-mini, Mistral Large 2, and LLaMA 3.1 70B) are particularly recommended for high-quality summaries and precise risk analyses.
\end{itemize}
This analysis shows that model selection strongly depends on the specific application requirements. The decision between proprietary and open-source models should be based on budget, quality requirements, and efficiency needs.

\subsection{Qualitative Evaluation}

\subsubsection{Evaluation with G-Eval}

\begin{table}[htbp]
\caption{Evaluation Metrics Using Reference Summaries (Sorted by Average)}
\vspace{3mm}
\centering
\small
\begin{tabular}{ccccccc}
\toprule
Evaluation Type & Coherence & Consistency & Fluency & Potential Impact & Relevance & $\varnothing$ \\
\midrule
Few-Shot Mistral 822B & 10 & 10 & 10 & 9 & 10 & 9.8 \\
Few-Shot GPT-4o mini & 10 & 10 & 10 & 9 & 10 & 9.8 \\
Zero-Shot Gemma2 9B & 10 & 10 & 10 & 9 & 10 & 9.8 \\
Zero-Shot GPT-4o mini & 10 & 10 & 10 & 9 & 10 & 9.8 \\
Few-Shot GPT-3.5 Turbo & 10 & 10 & 10 & 8 & 10 & 9.6 \\
Zero-Shot LLaMA3-2 90B & 10 & 10 & 9 & 9 & 10 & 9.6 \\
Zero-Shot Mistral 822B & 9 & 10 & 10 & 9 & 10 & 9.6 \\
Zero-Shot LLaMA3-1 8B & 10 & 10 & 10 & 8 & 10 & 9.6 \\
Zero-Shot Gemma 7B & 9 & 10 & 9 & 9 & 10 & 9.4 \\
Zero-Shot LLaMA3 70B & 9 & 10 & 10 & 8 & 10 & 9.4 \\
Zero-Shot LLaMA3-1 70B & 9 & 10 & 9 & 10 & 9 & 9.4 \\
Few-Shot LLaMA3 8B & 9 & 10 & 9 & 8 & 10 & 9.2 \\
Few-Shot Mistral 7B & 9 & 10 & 9 & 8 & 10 & 9.2 \\
Fine-Tuned GPT-4o mini 2024 & 8 & 10 & 9 & 9 & 10 & 9.2 \\
Fine-Tuned with GPT-3.5 & 8 & 10 & 9 & 9 & 10 & 9.2 \\
Zero-Shot GPT-4o & 9 & 10 & 10 & 9 & 8 & 9.2 \\
Zero-Shot Mistral 7B & 9 & 10 & 9 & 8 & 10 & 9.2 \\
Few-Shot LLaMA3 70B & 9 & 10 & 9 & 8 & 9 & 9.0 \\
Few-Shot Mistral Large 2 & 9 & 9 & 9 & 8 & 10 & 9.0 \\
Few-Shot LLaMA3-1 70B & 9 & 10 & 9 & 8 & 9 & 9.0 \\
Zero-Shot Mistral Large 2 & 9 & 10 & 9 & 8 & 9 & 9.0 \\
Few-Shot LLaMA3-2 90B & 9 & 10 & 9 & 8 & 8 & 8.8 \\
Few-Shot Gemma2 9B & 8 & 9 & 9 & 8 & 9 & 8.6 \\
Few-Shot LLaMA3-1 8B & 8 & 10 & 9 & 8 & 8 & 8.6 \\
Zero-Shot LLaMA3 8B & 8 & 8 & 9 & 8 & 10 & 8.6 \\
Zero-Shot GPT-3.5 Turbo & 9 & 8 & 9 & 8 & 8 & 8.4 \\
Few-Shot Gemma 7B & 8 & 0 & 8 & 8 & 0 & 4.8 \\
\bottomrule
\end{tabular}
\label{tab:geval}
\end{table}

The qualitative evaluation of models using G-Eval reveals interesting differences and similarities in terms of coherence, consistency, fluency, relevance, and potential impact. This analysis provides insights into the strengths of the models and their suitability for specific applications in supply chain management.
\begin{itemize}
    \item \textbf{Top-Performing Models}: Zero-Shot and Few-Shot GPT-4o mini, Zero-Shot Gemma2 9B, and Few-Shot Mistral 822B achieved the highest average score of 9.8. These models excel in generating coherent and consistent content, making them well-suited for applications where accuracy and precision are crucial.
\end{itemize}
\begin{itemize}
    \item \textbf{Potential Impact}: Some models received lower scores (8) in potential impact. This could indicate difficulties in precisely capturing the context of supply chain management, potentially limiting their practical applicability in decision-intensive areas.
\end{itemize}
\begin{itemize}
    \item \textbf{Few-Shot vs. Zero-Shot Configurations}: Interestingly, some Zero-Shot models (like Gemma2 9B and GPT-4o mini) performed as well as or better than their Few-Shot counterparts. This suggests these models can handle complex tasks well without specific training.
\end{itemize}
\begin{itemize}
    \item \textbf{Fine-Tuned Models}: While achieving high scores in coherence and consistency, fine-tuned models lagged behind the best Few-Shot and Zero-Shot models in potential impact and relevance. Fine-tuning may meet specific requirements but shows less flexibility and generalization ability in processing new content.
\end{itemize}
\begin{itemize}
    \item \textbf{Weaknesses of Individual Models}: The Few-Shot Gemma 7B model, with an average score of 4.8, showed clear weaknesses in coherence and relevance. These low values suggest limitations in architecture or training that could hinder reliable application.
\end{itemize}
\begin{itemize}
    \item \textbf{Open-Source Models and Consistency}: Open-source models like Gem\-ma2, Mistral 8x22B, Mistral 7B, and LLaMA consistently achieved high scores, often competing with proprietary models. This consistency demonstrates that open-source approaches can offer a valid alternative for high-quality, automated text generation.
\end{itemize}

\subsubsection{Human Evaluation}

To complement the quantitative analyses, a human evaluation was conducted to assess the quality of summaries generated by large language models (LLMs). Initially, the evaluation was designed to include 27 models with 10 articles each. However, due to the significant workload for participants, the methodology was refined to optimize efficiency while maintaining relevance. The final approach involved evaluating three articles per model. The models were selected based on their performance in quantitative metrics (ROUGE, BLEU, and BERTScore) and vendor diversity to ensure balanced representation.

The evaluation focused on five key criteria: coherence, consistency, fluency, potential impact, and relevance. Feedback was collected from 31 participants over a month-long period, providing valuable insights into the strengths and limitations of the models. The participant group consisted of 11 men (35.5\%) and 20 women (64.5\%), with an average age of 29 years. The study took place from October 1 to October 30, 2024.

The eight models selected for evaluation were: Zero-shot GPT-4o, Few-shot GPT-4o mini, Zero-shot Mistral Large 2, Few-shot Open-Mistral-8x22b, Few-shot LLaMA 3.1 70B, Zero-shot LLaMA 3 70B, Few-shot Gemma2-9b, and Fine-Tuned GPT-3.5.

A systematic analysis of the results was conducted, compiling participant ratings into a table (see Table \ref{tab:human_eval}). For each model, average scores were calculated across all evaluation criteria, with an overall average determined from the three assessed articles. These results were then compared with G-Eval ratings and quantitative metrics to derive meaningful insights.

\begin{table}[htbp]
\caption{Statistical Analysis of the Human Evaluation (Sorted by Average)}
\vspace{3mm}
\label{tab:human_eval}
\centering
\small
\resizebox{\columnwidth}{!}{
\begin{tabular}{lcccccc}
\toprule
Evaluation Type & Coherence & Consistency & Fluency & Potential Impact & Relevance & $\varnothing$ \\
\midrule
Few-Shot Gemma2 9B       & 6.91 & 6.67 & 7.28 & 7.17 & 6.59 & 6.92 \\
Zero-Shot Mistral Large 2   & 6.96 & 6.70 & 7.35 & 6.88 & 6.58 & 6.89 \\
Few-Shot GPT-4o mini     & 6.92 & 6.54 & 7.24 & 6.99 & 6.70 & 6.88 \\
Few-Shot Mistral 822B    & 6.89 & 6.40 & 7.29 & 7.08 & 6.42 & 6.82 \\
Zero-Shot GPT-4o         & 6.88 & 6.51 & 7.27 & 6.74 & 6.46 & 6.77 \\
Zero-Shot LLaMA3 70B     & 6.56 & 6.30 & 6.99 & 7.02 & 6.30 & 6.63 \\
Few-Shot LLaMA3-1 70B    & 6.52 & 6.11 & 7.03 & 7.03 & 6.29 & 6.60 \\
Fine-Tuned with GPT-3.5 & 6.43 & 6.13 & 6.91 & 6.35 & 5.91 & 6.35 \\
\bottomrule
\end{tabular}
}
\end{table}

\paragraph{Key Findings}
\begin{itemize}
    \item \textbf{Coherence and Consistency}: Discrepancies were observed between human evaluation and G-Eval. For instance, Few-Shot Gemma2 9B ranked first in human evaluation despite its lower rank (23) in G-Eval. This suggests that human evaluators prioritize narrative structure and natural flow, which may not be fully captured by algorithmic metrics.

    \item \textbf{Fluency and Potential Impact}: Human evaluators consistently rated fluency lower than G-Eval, indicating that LLMs may not fully meet human expectations for natural expression and readability. This highlights the need for model optimization to improve language patterns and persuasiveness.

    \item \textbf{Relevance}: Models like Few-Shot Gemma2 9B and Zero-Shot Mistral Large 2 were highly rated for relevance, suggesting they provide useful information tailored to reader needs, even if their formal G-Eval scores were lower.

    \item \textbf{Discrepancy Between Human Evaluation and G-Eval}: Human ratings were consistently lower than G-Eval scores, indicating that human evaluators are more critical and may capture nuances that automated systems overlook.

    \item \textbf{Performance of Open-Source Models}: Open models such as Gem\-ma2 9B and Mistral 8x22B performed well in human evaluation, demonstrating their potential for practical applications.

    \item \textbf{Influence of Prompting Approach}: No consistent superiority was observed between Few-Shot and Zero-Shot configurations in human evaluation, contrasting with G-Eval, where Few-Shot models generally performed better.
\end{itemize}

%\paragraph{Challenges and Limitations.}
The human evaluation process faced several challenges. Participants reported fatigue due to the volume of text, leading some to abandon the evaluation or complete it over multiple days. Additionally, the technical language and sophisticated style of the articles posed barriers for some evaluators. Furthermore, the models exhibited a tendency to generalize, often replacing specific place names with higher-level regions, which resulted in inaccuracies in location information.

%\paragraph{Conclusion}
The results of the human evaluation underscore the importance of complementing automated metrics with human judgment. The discrepancies between human and G-Eval assessments highlight the need for further research to develop evaluation methods that better align with human perceptions. This study emphasizes the value of human evaluation in refining LLMs for more precise and informative news analysis, particularly in applications requiring high readability and relevance.

\section{Discussion}
\subsection{Answering the Research Questions}

The study addressed several key research questions, yielding the following insights:
\begin{itemize}
    \item \textbf{Suitability for News Summarization in Risk Analysis}: Modern LLMs, particularly GPT-4o mini, GPT-4o, and Mistral Large 2, show strong effectiveness for automated risk analysis. Statistical analyses confirm their consistent high performance across multiple metrics, including coherence, relevance, and fluency.
    \item \textbf{Summary Quality}: The models exhibit strong readability and coherence, with their responses implicitly incorporating duplicate detection and risk identification. This validates their ability to generate high-quality summaries tailored to supply chain risk analysis.
    \item \textbf{Comparison of Zero-Shot, Few-Shot, and Fine-Tuning}: Zero-shot configurations prove highly efficient when using powerful models, while few-shot approaches often enhance qualitative outcomes. Few-shot prompting, especially with GPT-4o mini, demonstrated strong performance. Fine-tuning yielded mixed results, suggesting that its effectiveness depends on the specific context.
\end{itemize}

The study confirms the potential of modern LLMs for automated news summarization in supply chain risk analysis. Few-shot GPT-4o mini excels across multiple evaluation dimensions. The discrepancy between automated metrics and human evaluation underscores the necessity of combined quantitative and qualitative assessments.

\subsection{Practical Implications}

The findings have significant practical implications for organizations leveraging LLMs in supply chain risk analysis:
\begin{itemize}
    \item \textbf{Efficiency Gains}: Models like Few-Shot GPT-4o mini, Zero-Shot GPT-4o, and Zero-Shot Mistral Large 2 accelerate risk monitoring through rapid analysis and summarization, enabling timely decision-making.
    \item \textbf{Cost Efficiency}: Budget-friendly models such as Gemma2 9B (Few-Shot), Llama 3 70B (Zero-Shot), and Mistral 822B (Few-Shot) offer strong performance for resource-constrained organizations, making advanced AI tools more accessible.
    \item \textbf{Real-Time Capability}: LLaMA models’ high processing speeds enable real-time analysis, critical for proactive risk management and dynamic supply chain environments.
\end{itemize}

\subsection{Model-Specific Recommendations}

Based on the evaluation results, the following model-specific recommendations are proposed:
\begin{itemize}
    \item \textbf{GPT-4o mini (Few-Shot)}: The top performer in both human and quantitative evaluations, offering good cost-efficiency. Regression analysis confirms that it consistently maintains low latency and cost while delivering strong performance. This makes it an ideal choice for organizations seeking an optimal balance between accuracy, speed, and resource utilization.
    \item \textbf{GPT-4o (Zero-Shot)}: Exceptional quantitative performance, but at a higher cost. Its strong accuracy makes it suitable for high-stakes applications where precision is critical.
    \item \textbf{Mistral Large 2 (Zero-Shot)}: Correlation analysis indicates that Mistral Large 2 (Zero-Shot) achieves an effective balance between cost and performance. This model is recommended for organizations seeking a reliable, yet cost-effective solution.
    \item \textbf{LLaMA 3.1 70B (Few-Shot)}: Optimal performance-efficiency balance, ideal for time-sensitive applications requiring rapid analysis.
    \item \textbf{Gemma2 9B (Few-Shot)}: Strong human evaluation results and cost-efficiency, suitable for budget-limited projects or smaller-scale implementations.
\end{itemize}

\section{Summary and Conclusion}

This study comprehensively investigated the potential of modern large language models (LLMs) for automated news summarization in supply chain risk analysis. The central research questions aimed to evaluate the models' ability to accurately summarize relevant content and analyze their performance in terms of readability, duplicate detection, and risk identification. The results demonstrate that LLMs provide valuable support for risk analysis by enabling companies to efficiently identify critical news content and respond swiftly to potential disruptions in their supply chains.

Quantitative and qualitative analyses revealed that the Few-Shot GPT-4o mini model delivered outstanding performance, excelling in both automated metrics and human evaluations while offering excellent cost-effectiveness. Models like Zero-Shot GPT-4o and Mistral Large 2 also demonstrated impressive results across various evaluation dimensions. Few-Shot GPT-4o mini, in particular, showed exceptional performance.

A key insight from this study is the importance of a combined evaluation methodology. Integrating quantitative metrics, LLM-based evaluators like G-Eval, and human assessments provided a comprehensive and nuanced understanding of model performance. This multidimensional evaluation also revealed discrepancies between automated and human assessments, underscoring the necessity of holistic evaluation approaches.

The results highlight that LLMs, through targeted prompting techniques such as Zero-Shot and Few-Shot, can generate precise and consistent summaries. These summaries form a robust decision-making foundation for risk managers in global supply chains. At the same time, the study emphasized challenges related to model biases, factual accuracy, and ethical considerations.

Regression analysis identified total costs and output speed as the primary factors influencing latency, highlighting the importance of balancing performance and efficiency. These findings are particularly relevant for corporate settings, where practical implementation requires optimizing these trade-offs. From a business perspective, this study provides companies with actionable insights for selecting and adapting LLMs based on specific needs. The results emphasize that model choice should consider factors such as accuracy, speed, and cost.

\subsection{Limitations}
While the study provides valuable insights, several limitations must be acknowledged:
\begin{itemize}
    \item \textbf{Dataset Constraints}: The limited dataset size may affect the generalizability of the results. Future work should incorporate larger, more diverse datasets to validate findings across different contexts.
    \item \textbf{Model Bias}: Potential biases in LLMs may impact objectivity, requiring deeper investigation \cite{zhang2024benchmarking}. Addressing these biases is critical for ensuring fair and accurate risk analysis.
    \item \textbf{Long-Term Performance}: The stability of model performance over time remains unverified. Longitudinal studies are needed to assess how models adapt to evolving risks and data patterns.
    \item \textbf{Fine-Tuning Limitations}: Resource constraints prevented evaluation of alternative models like LLaMA 3.1/3.2. Future research should explore fine-tuning with a broader range of models and datasets.
\end{itemize}

\subsection{Conclusion and Outlook}
In summary, LLMs like GPT-4o mini and Mistral Large 2 represent promising tools for automated news summarization in risk management. However, the selection of models should align with the specific requirements for accuracy, speed, and cost efficiency. This study contributes significantly to the scientific discourse on AI in supply chain management and establishes a solid foundation for future research and applications in supply chain risk analysis.

Future research should explore advanced fine-tuning techniques, domain-specific customizations, and diverse datasets to enhance model robustness and reliability. Expanding experimental studies to a wider range of news sources and risk scenarios will strengthen the generalizability of the findings. Furthermore, a deeper investigation of the ethical implications of automated AI systems, along with the development of clear guidelines for large language model (LLM) applications in risk management, will help foster corporate confidence in adopting these technologies.

By addressing these limitations and following the research directions described, the field can advance toward more reliable, efficient, and ethical AI-driven solutions for supply chain risk analysis, ultimately enabling organizations to navigate complex global challenges with greater agility, accuracy, and resilience.

\bibliographystyle{splncs04}
\bibliography{main}

\end{document}